\documentclass[sn-mathphys,Numbered]{sn-jnl}

\usepackage{graphicx}%
\usepackage{multirow}%
\usepackage{amsmath,amssymb,amsfonts}%
\usepackage{amsthm}%
\usepackage{mathrsfs}%
\usepackage[title]{appendix}%
\usepackage{xcolor}%
\usepackage{textcomp}%
\usepackage{manyfoot}%
\usepackage{booktabs}%
\usepackage{algorithm}%
\usepackage{algorithmicx}%
\usepackage{algpseudocode}%
\usepackage{listings}%

\theoremstyle{thmstyleone}%
\theoremstyle{thmstyletwo}%

\theoremstyle{thmstylethree}%

\begin{document}

\title[Sentiment Analysis for AI vs. Human made Texts]{Sentiment analysis and random forest to classify LLM versus human source applied to Scientific Texts}


\author*[1]{\fnm{Javier J.} \sur{Sanchez-Medina}}\email{javier.sanchez@ulpgc.es}

\affil*[1]{\orgdiv{CICEI - IUCES}, \orgname{ULPGC}, \orgaddress{\street{Campus de Tafira}, \city{Las Palmas de Gran Canaria}, \postcode{35017}, \state{Islas Canarias}, \country{Spain}}}

\abstract{After the launch of \emph{ChatGPT v.4} there has been a global vivid discussion on the ability of this artificial intelligence powered platform and some other similar ones for the automatic production of all kinds of texts, including scientific and technical texts. This has triggered a reflection in many institutions on whether education and academic procedures should be adapted to the fact that in future many texts we read will not be written by humans (students, scholars, etc.), at least, not entirely. 

In this work it is proposed a new methodology to classify texts coming from an automatic text production engine or a human, based on Sentiment Analysis as a source for feature engineering independent variables and then train with them a \emph{Random Forest} classification algorithm. 

Using four different sentiment lexicons, a number of new features where produced, and then fed to a machine learning random forest methodology, to train such a model. 

Results seem very convincing that this may be a promising research line to detect fraud, in such environments where human are supposed to be the source of texts.}




\maketitle

\section{Introduction}\label{sec1}

For developing the present paper, a potential author reading the journal instructions may find a new specification. Nowadays is common to read sentences like the following: "Large Language Models (LLMs), such as ChatGPT, do not currently satisfy our authorship criteria". Large language models (LLMs) are language models capable of "general-purpuse" language understanding and generation. LLMs are trained using massive volumes of data and using massive computational facilities (\cite{radford2019better}).

For every computer scientist it is very well known the Turing Test, or the Imitation Game, as Alan Turing called it, as a milestone where we could acknowledge that Artificial Intelligence is really here. In 1950 \cite{turing2009computing} Turing proposed that game where when the human in one end of a teletype line could not distinguished if the entity to which he or she was communicating was another human or a machine, was the Rubicon beyond we could safely say Artificial Intelligence was here.

In \cite{uchendu2021turingbench}, the Turing Test is carried out with a number of models, namely \textit{GPT-1, GPT-2\_small, GPT-2\_medium, GPT-2\_large, GPT-2\_xl, GPT-2\_PyTorch, GPT-3, GROVER\_base, GROVER\_large, GROVER\_mega, CTRL, XLM, XLNET\_base, XLNET\_large, FAIR\_wmt19, FAIR\_wmt20, TRANSFORMER\_XL, PPLM\_distil} and \textit{PPLM\_gpt2}. All of them tryed agains human generated samples. Their results shown that the most human-like indistinguishable texts, yielding lowest F1 scores where FAIR\_wmt20 and GPT-3. 

The Generative Pre-trained Transformer (GPT, \cite{floridi2020gpt}), one the the best known LLMs, is an autoregressive language model including deep learning under the hood, able to produce credible texts. Last March 14, 2023, the Generative Pre-trained Transformer 4 (GPT-4, \cite{openai2023gpt4}) was announced to be available, triggering a social earthquake for the consequences of the evolution of this technology. Every industry and institution where there is the need and assumption that texts are human made has been affected by the recent news on artificial intelligence based platforms that are able to produce from a web page html code to a poem in Latin. 

The social, industrial, cultural and academic impact of the coming of this new technology is believed to be deep (\cite{gozalo2023chatgpt}). For example, in \cite{biswas2023role} it is shown how ChatGPT can provide to public heath issues, mainly as precise and interactive information deliverer.


Academic institutions are particularly aware of the importance of this technological leap. For example, producing a scientific paper or a technical report on the topic one may think of, is a matter of just seconds using this platforms (\cite{firat2023chat}). In this context and many others, and in this time of adaptation to the new technology it may be useful to have tools that discriminate texts being produced by human or by software.

A positive approach to this new tool can be found too. For example, in \cite{emenike2023title} is discussed what the impact of students using ChatGPT for their assignments preparation may be, in particular around Chemistry teaching. The potential benefits and limitations of embracing that new technology into the education system is explored and discussed.

In the present work a new methodology is proposed in order to help out in that effort. A classic machine learning algorithm is used (Random Forest) to detect whether a text was human or ChatGPTv.3.5 produced. The key of this methodology is that the features that support the model trained are produced after applying Sentiment Analysis to such texts, using of four  standard lexicon, namely "afinn"(\cite{nielsen2011new}, "bing"(\cite{hu2004mining}, "nrc" (\cite{mohammad2013nrc}), or "loughran" (\cite{loughran2020textual}. Also, data augmentation has been used, to extend the dataset and to achieve better evaluation scores.

The rest of this paper is organized as follows: In section \ref{SoA} a few related works are commented. Then it is shared the methodology proposed, starting for the Data Ingestion and Preparation (\ref{Dataprep}. In section \ref{FeatureEngineering} the pre-processing and development of the needed explanatory (independent) features is shared. A short description of the main learning model methodology used, Random Forests (\ref{randomforest} is followed by the learning methodology description, in section \ref{methodology}. Results and some concluding remarks are shared in section \ref{results} and \ref{conclusions}.

\section{State of the Art}\label{SoA}
In this section there are some related works commented. Literature is not abundant (yet) in this specific topic.

In \cite{jawahar2020automatic}, a research paper previous to the publication of ChatGPT, the criticality of having detectors that discriminate whether a texts was human made or AI made. In that work, they propose 5 different important research lines that seem still valid to the date of the publication of this work:
\begin{enumerate}
    \item Leveraging auxiliary signals
    \item Assessing veracity of the text
    \item Building generalizable detectors
    \item Building interpretable detectors
    \item Building detectors robust to adversarial attacks
\end{enumerate}

In that work, one can see that it is mostly about detectors, and that, in the end to these authors, creating methodologies for the detection of "fake", artificially generated texts, is of upmost importance.


In \cite{gao2022comparing} we have the closer to the presented methodology work found. They compare 50 abstracts coming from medical journals to 50 abstracts generated by ChatGPT, based on their titles and journals. A three folded detection method is used, including an output detector, plagiarism detector and human reviewers. Results are very promising, but this methodology has the need of human in the loop, which may be seen as a severe limitation in order to setup a system to automatically and robustly classify texts as human made or LLM made. 

Also, since only 8\% of ChatGPT generated texts correctly followed the specific journal’s formatting requirements, that bias makes it quite easy to detect which category every text falls in.

In the presented methodology in this paper, since the length and formatting of texts are summarized into the Sentiment Analysis based obtained features, there is no possible information leaking between categories, making our proposed method more robust, format independent in that sense, even if results in such paper seems very good.

In \cite{wu2023llmdet} authors used the perplexity metric to determine what source is more likely to be the source of a text. That means comparing the perplexity metric for each of a number of models, and to a human model, and then, the lower the perplexity, the more likely that model is the source of the text. 

This research approaches the current problem in a very interesting way, but a limitation of it would be that it relies on accurate \textit{updated} models of each LLMs, and of human made texts. That does not seem trivial to have. Regarding human texts alone, there is so much diversity on how humans of different countries write (using the same official language), that it seems challenging to produce such good results invariantly. Also, LLMs are in constant evolution, so this approach also may be limited as they will most likely tend to mimic the probability distributions of human made texts, as long as they continue learning from them. 

There are some works based on statistical approaches to detect LLMs sources in texts, like \cite{gehrmann-etal-2019-gltr} and \cite{frohling2021feature}. Results are good, but with older LLMs. In this present work we proposed a new approach, based on Sentiment Analysis, that could well complement feature based training approaches like in \cite{frohling2021feature}. 

A recent approach to this problem is to tweak LLMs in order they produce "watermarks" in the produced texts, to allow detectors to find them and determined whether a text was produced by a LLM and which one. In \cite{kirchenbauer2023watermark} a method in this line is presented. The idea is very interesting but it would need the cooperation of LLMs programmers/owners, but it seems like the goal of each LLM is exactly against that spirit.

\section{Data Ingestion and Preparation. Exploratory Analysis}\label{Dataprep}

In this section it is shared how the data used to test this new methodology has been ingested, and prepared, before training. The goal was to have two sets of instances coming from Human and ChatGPT as generators, to then develop a methodology to automatically detect which is which. We have selected 68 recent papers in the journal "New Phytologist", edited by Wiley and Sons, and recorded titles and abstracts. Then we launched 68 queries to the interactive chat of ChatGPT (v3), in order to make it produce an equivalent abstract. For example, if the title of one paper in "New Phytologist" was titled \emph{"Delay of Iris flower senescence by protease inhibitors"}, the query asked was \emph{"Write a text on the Delay of Iris flower senescence by protease inhibitors, shorter than 1500 characters, white spaces included"}. That query produced a text that was stored and tagged as produced by ChatGPT.

\begin{figure}[!tbp]
  \centering
  \begin{minipage}[b]{0.475\textwidth}\label{35_most_frequent_stemmed_words_NewPhytologist}
    \includegraphics[width=\textwidth]{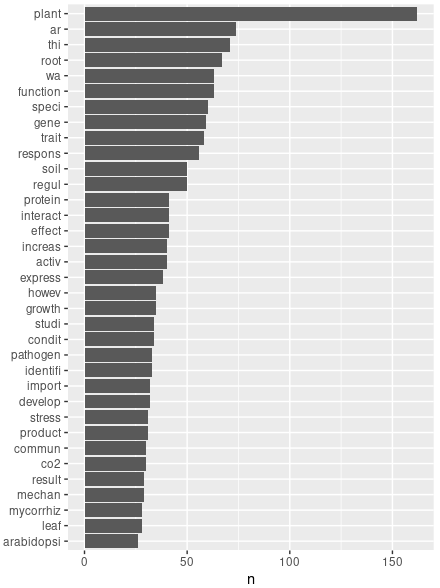}
    \caption{35 most frequent stemmed words in the original paper abstracts, as published}
  \end{minipage}
  \hfill
  \begin{minipage}[b]{0.475\textwidth}\label{35_most_frequent_stemmed_words_ChatGPT}
    \includegraphics[width=\textwidth]{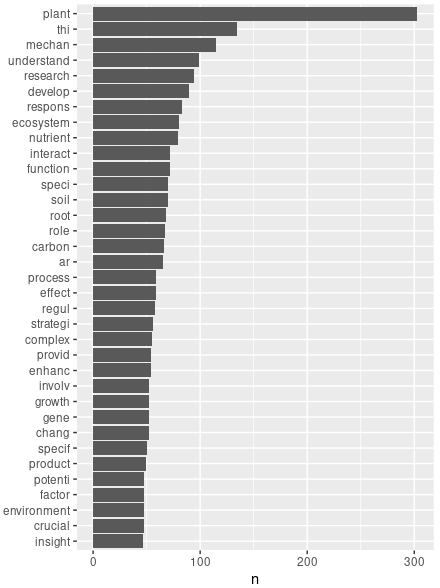}
    \caption{35 most frequent stemmed words in the ChatGPT v.3 generated paper abstracts}
  \end{minipage}
\end{figure}

Then all texts are cleaned of stop words, and a stemming process is applied, using the R libraries \emph{"tidytext"} and \emph{"SnowballC"}. In figures \ref{35_most_frequent_stemmed_words_NewPhytologist} and \ref{35_most_frequent_stemmed_words_ChatGPT} is represented the 35 most frequent stemmed words for each case.  

\begin{figure}[!tbp]
  \centering
    \begin{minipage}[b]{0.95\textwidth}\label{Frequencies_ChatGPT_vs_NewPhytologist}
        \includegraphics[width=\textwidth]{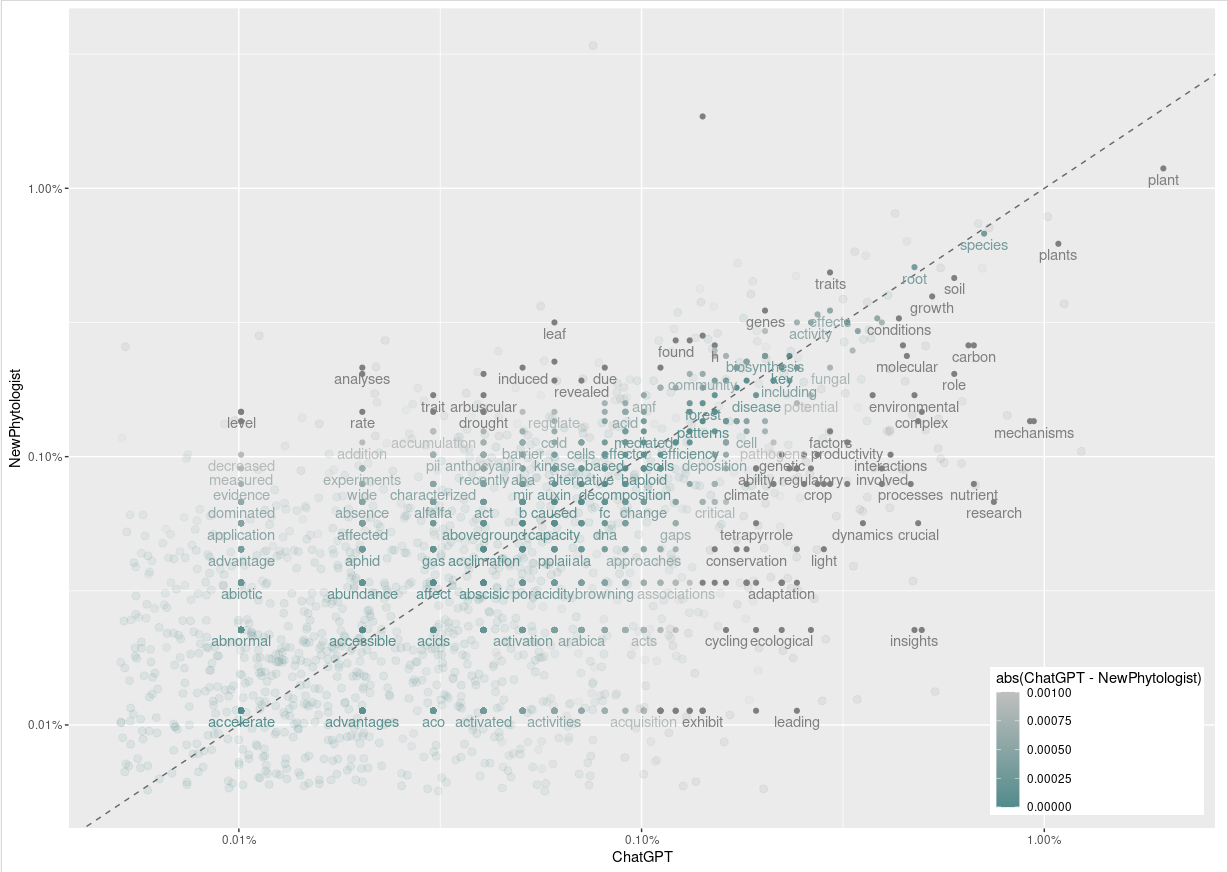}
        \caption{Stemmed words frequencies in the original papers versus the ChatGPT generated}
    \end{minipage}
\end{figure}

As can be seen in figure \ref{Frequencies_ChatGPT_vs_NewPhytologist}, there is no evident linear correlation between the frequencies of the stemmed words obtained from both sets of texts. The 95\% confidence interval of the Pearson's product-moment correlation ranges from 0.5508709 to 0.6221319, averaging 0.5876398, which is also indicating no linear correlation between the two sets. 

\begin{figure}[!tbp]
  \centering
    \begin{minipage}[b]{0.95\textwidth}\label{Inner_join_words_lexicons}
        \includegraphics[width=\textwidth]{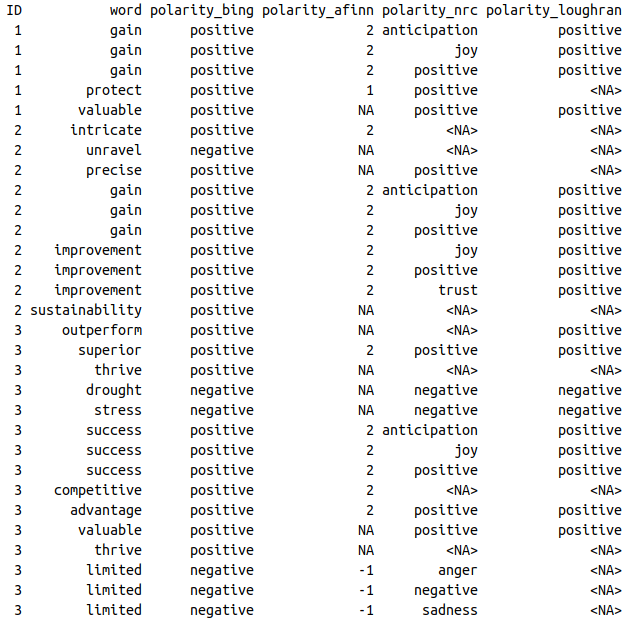}
        \caption{Inner Join between texts words and lexicons}
    \end{minipage}
\end{figure}

\section{Feature Engineering. Data Imputation, Data Augmentation and Sentiment Analysis}\label{sentiment_analysis}

The core of the proposed methodology is using sentiment analysis derived attributes as explanatory variables for a predictive modeling of texts being original from the journal or ChatGPT generated. The goal is to provide a reproducible simple methodology to build detectors that can be applied to all fields, just by applying sentiment analysis using publicly available lexicons. 

It has been used four different Lexicons namely "Bing" (\cite{hu2004mining}), "Afinn" (\cite{aung2017sentiment}), "Nrc" (\cite{mohammad2013nrc} and "Loughran-McDonald" (\cite{loughran2020textual}. These are 

The "Bing" lexicon is binary in the sense that every entry there is assigned just a "positive" or "negative" polarity. 

The "Afinn" lexicon however assigns a sentiment valence to each included word with an integer between -5 (negative) and +5 (positive). 

The "Nrc" lexicon assigns each word one of 10 different categories including eight basic emotions ("anger", "fear", "anticipation", "trust", "surprise", "sadness", "joy" and "disgust") and two sentiments ("negative" and "positive"). 

Finally, the "Loughran-McDonald" lexicon assigns one of 6 sentiment indicators, including "positive", "negative", "constraining", "litigious", "superfluous" and "uncertainty".

\subsection{Preprocessing}
In this subsection it will be described the data preprocess applied. Using the \emph{tidytext} R library, we have filtered out the stop words from all texts. Then, also using the same library it have loaded the four lexicons employed, in order to assign sentiment values to all the words in each document. 

In figure \ref{Inner_join_words_lexicons} it is represented how the "inner join" of the 4 lexicons and 30 rows of the words included in each document are combined. Words with no match in any of the lexicons are removed. 

In that table we have for each document (every different ID) the inner join of 

\subsection{Feature Engineering}\label{FeatureEngineering}
A \emph{dataset} with a row per document was needed. Therefore, the aggregation of all the pairs ('word'-'sentiment'). 
The number of cleaned words per document are counted. Then, the number of positive words 

\emph{Bing}: The ratio between the number of positive and negative polarity words in every document, over the number of words was calculated.

\emph{Afinn}: The average and standard deviation valence of all words that hit a word in the \emph{Afinn} lexicon is calculated. Some documents did not yield any word included in the \emph{Afinn} lexicon. That only happened for the \emph{Afinn} dictionary and for that reasing, there is a later data imputation phase in the proposed methodology.

\emph{Loughran-McDonald}: Only "positive", "negative", "constraining" and "uncertainty" sentiments were found in the texts preprocessed. For them the ratio between the number of words tagged with each one of the categories, over the number of words of each document was calculated.

\emph{Nrc}: 
("anger", "fear", "anticipation", "trust", "surprise", "sadness", "joy" and "disgust") and two sentiments ("negative" and "positive").

From that raw data it has been developed a number of features before starting training the classification model. First, counting the number of negative and positive "Bing" polarity words by abstract, that is two new columns. 
\begin{figure}[!h]
  \centering
    \begin{minipage}[b]{0.95\textwidth}\label{Data_str_imputed_complete_randomized}
        \includegraphics[width=\textwidth]{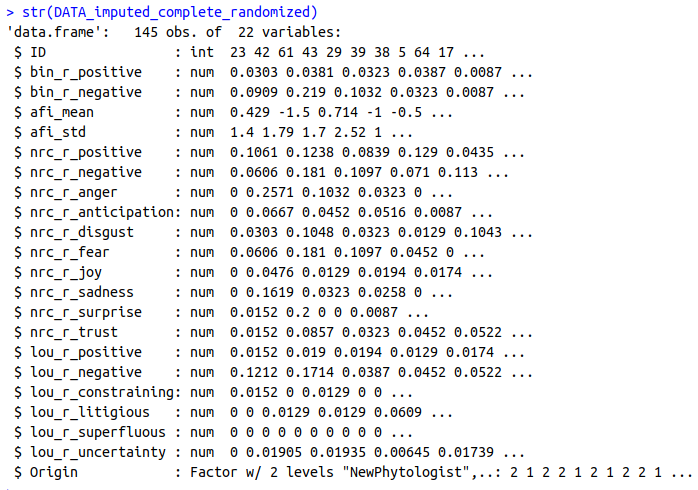}
        \caption{Data Prepared Structure}
    \end{minipage}
\end{figure}

\begin{figure}[!h]
  \centering
    \begin{minipage}[b]{0.95\textwidth}\label{Data_summary_imputed_complete_randomized}
        \includegraphics[width=\textwidth]{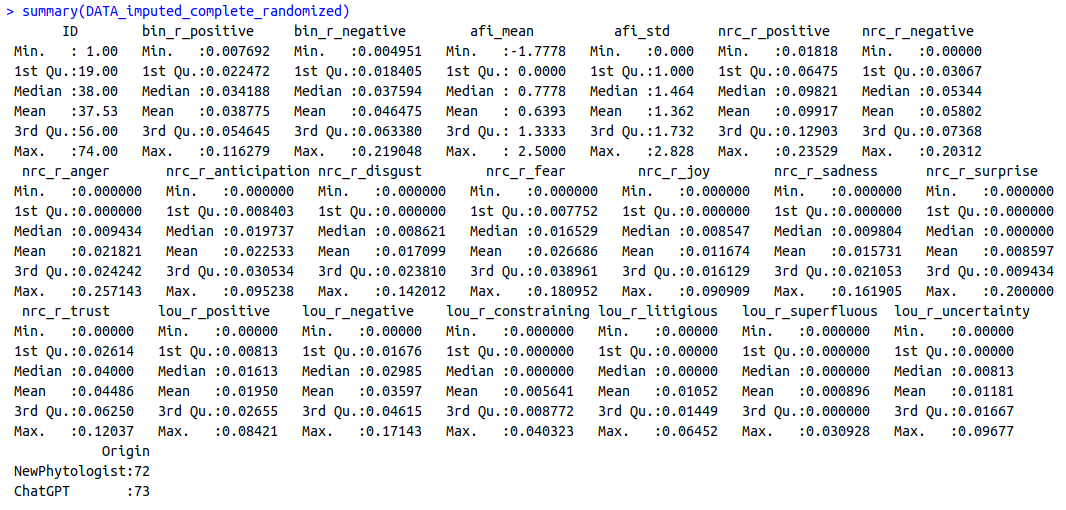}
        \caption{Data Prepared Summary}
    \end{minipage}
\end{figure}

\section{Random Forest}\label{randomforest}
Random forests is probably the most used ensemble methodology for its ability to combine the instances variance modeling while avoiding over-fitting. As published in \cite{Breiman2001}, it is an ensemble classification model learning methodology based on training a large number of decision trees, and then, for classification tasks, the most frequent detected category is chosen. Simple and effective, leads also to interpretable models, through techniques like Explainable Matrix (ExMatrix)(\cite{9222255}), which adds value to the trained results.

\section{Methodology}\label{methodology}

One prepared the data, a Random Forest Ensemble was trained, in order to detect whether a text was obtained from the "New Phytologist" journal or ChatGPT. Since, after cleaning, there were only 72 and 73 instances in each category, a stratified 10-folded cross validation methodology was used to train and evaluate the model. In a future study a more extensive dataset will be used, and probably data augmentation will also be considered.

Using the Waikato University WEKA 3 java based library (\cite{frank2016weka}), a Random Forest model was trained. "RandomForest" constructs random forests by bagging ensembles of random trees. For this publication 100000 trees were used. Other important setting arguments were the minimum number of instances per leaf (M), which was set to 1; and the minimum numeric class variance proportion of train variance for split, which as set to 1e-3 (default value).

The training execution time was 1.48 seconds, in a PC, running Ubuntu 22.04.3 LTS with the following characteristics:
\begin{itemize}
    \item Hardware model: Gigabyte Technology Co., Ltd. B560M DS3H V2
    \item Memory: 16 GiB
    \item Processor: 11th Gen Intel® Core™ i7-11700 @ 2.50GHz × 16
\end{itemize}





\section{Results}\label{results}
In this section the results are summarized. 
\begin{table}[h]
\caption{Results Summary}\label{results_summary}%
\begin{tabular}{@{}lll@{}}
\toprule
Statistic & Abs. value  & Percentage\\
\midrule
Correctly Classified Instances    & 122  & 84.1379 \%  \\
Incorrectly Classified Instances    & 23   & 15.8621 \%  \\
Kappa statistic    & 0.6827   & \\
Mean absolute error & 0.3016 & \\
Root mean squared error & 0.3724 & \\
Relative absolute error & & 60.3095 \% \\
Root relative squared error  & & 74.4571 \% \\
Total Number of Instances & 145 & \\
\botrule
\end{tabular}
\footnotetext{Source: This is an example of table footnote. This is an example of table footnote.}
\end{table}

 \begin{table}[h]
\caption{Detailed Accuracy By Class}\label{detailed_accuracy}%
\begin{tabular}{@{}llll|l@{}}
\toprule
TP Rate & FP Rate & Precision & Recall & Class\\
\midrule   
0.849 & 0.167 & 0.838 & 0.849 & ChatGPT \\
0.833 & 0.151 & 0.845 & 0.833 & NewPhytologist\\
0.841 & 0.159 & 0.841 & 0.841 & Weighted Avg. \\   
\toprule
F-Measure & MCC & ROC Area & PRC Area & Class\\
\midrule   
0.844 & 0.683 & 0.880 & 0.870 & ChatGPT\\
0.839 & 0.683 & 0.880 & 0.885 & NewPhytologist\\
0.841 & 0.683 & 0.880 & 0.877 & Weighted Avg.\\ 
\botrule
\end{tabular}
\footnotetext{Source: This is an example of table footnote. This is an example of table footnote.}
\end{table}

 \begin{table}[h]
\caption{Confusion Matrix}\label{confusion_matrix}%
\begin{tabular}{@{}ll|l@{}}
\toprule
a & b & classified as\\
\midrule 
 62 & 11 & a = ChatGPT\\
 12 & 60 & b = NewPhytologist\\
\botrule
\end{tabular}
\footnotetext{Source: This is an example of table footnote. This is an example of table footnote.}
\end{table}

As can be seen in table \ref{results_summary}, accuracy is 0.84, which seems a solid result, for this first round of a methodology. According to Cohen (\cite{cohen1960coefficient}), kappa value being larger than .6 suggests a substantial agreement (or a residual probability of results being by chance). 

Other  summarized parameters seem very nice, like Root Mean Squared Error, which is considered a good value under 0.75 and the obtained is 0.37.

Looking now at table \ref{detailed_accuracy} once can see that the trained model works quite similarly for both classes, obtaining solid good results for FP Rate 

Also, the Area under ROC seems very good, with a value closer to 0.9 than to 0.8, showing a very good performance of the trained classification model at all classification thresholds. The F-Measure also shows a great balance between Precision and Recall, stable across the two categories considered.

All in all, results are very promising that this is a good way to go regarding building models to predict whether texts are created by humans or AI. 

\section{Conclusions}\label{conclusions}
In this paper it is shown that a new methodology, based on sentiment analysis feature engineering is possible to train models that can detect, with a acceptable accuracy when a text is produced by an LLM tool or a human. 

Further research is needed: There is GPT-4, and other equivalent tools, where it should be needed to repeat a similar methodology in order to see if it is still wise to use sentiment analysis based feature engineering, to classify texts. 

Also, the combination of Sentiment Analysis features with other feature based approaches, statistic based methodologies, for instance, seems a very promising way to go.


\bibliography{sn-bibliography}

\end{document}